\documentclass{article}

\usepackage{arxiv}

\usepackage{cite}

\newcommand{\citep}[1]{\cite{#1}}
\newcommand{\citet}[1]{\cite{#1}}

\usepackage{amsmath,amssymb,amsthm}
\usepackage{dsfont,mathtools}
\usepackage{xspace}
\usepackage{epsfig}
\usepackage{algorithm}
\usepackage[noend]{algorithmic}
\usepackage{color}
\usepackage{url}
\usepackage{graphicx}
\usepackage{tikz}
\usepackage{float}
\usepackage{booktabs}
\usepackage{makecell}
\usepackage{multicol}
\usepackage{multirow}
\usepackage{csquotes}
\usepackage{wasysym}
\usepackage{longtable}
\usepackage{colortbl}
\usepackage{paralist}

\definecolor{colorEAXR}{RGB}{27,147,108}
\definecolor{colorEAXRGPX}{RGB}{212,85,4}
\definecolor{colorLKHRIPT}{RGB}{105,100,170}

\title{Anytime Behavior of Inexact TSP Solvers and Perspectives for Automated Algorithm Selection}

\author{
  Jakob Bossek \\
  Optimisation and Logistics\\
  The University of Adelaide\\
  Adelaide, Australia \\
  \texttt{jakob.bossek@adelaide.edu.au} \\
   \And
  Pascal Kerschke \\
  Information Systems and Statistics \\
  University of M{\"u}nster \\
  M{\"u}nster, Germany \\
  \texttt{kerschke@uni-muenster.de}
  \And
  Heike Trautmann \\
  Information Systems and Statistics \\
  University of M{\"u}nster \\
  M{\"u}nster, Germany \\
  \texttt{trautmann@wi.uni-muenster.de}
}

\begin{document}
\maketitle

\pagestyle{plain}
\thispagestyle{fancy}
\lfoot{\vspace*{-1.25cm}\rule{\columnwidth}{0.2pt}\\\footnotesize \textcopyright2020 IEEE. Personal use of this material is permitted. Permission from IEEE must be obtained for all other uses, in any current or future media, including reprinting/republishing this material for advertising or promotional purposes, creating new collective works, for resale or redistribution to servers or lists, or reuse of any copyrighted component of this work in other works.\\ This version has been accepted for publication at the \textit{IEEE Congress on Evolutionary Computation (IEEE CEC)} 2020, which is part of the \textit{IEEE World Congress on Computational Intelligence (IEEE WCCI)} 2020.}\cfoot{}

\begin{abstract}
The Traveling-Salesperson-Problem (TSP) is arguably one of the best-known $\mathcal{NP}$-hard combinatorial optimization problems. The two sophisticated heuristic solvers LKH and EAX and respective (restart) variants manage to calculate close-to optimal or even optimal solutions, also for large instances with several thousand nodes in reasonable time. In this work we extend existing benchmarking studies by addressing anytime behaviour of inexact TSP solvers based on empirical runtime distributions leading to an increased understanding of solver behaviour and the respective  relation to problem hardness.
It turns out that performance ranking of solvers is highly dependent on the focused approximation quality. Insights on intersection points of performances offer huge potential for the construction of hybridized solvers depending on instance features. Moreover, instance features tailored to anytime performance and corresponding performance indicators will highly improve automated algorithm selection models by including comprehensive information on solver quality.
\end{abstract}


\section{Introduction}

The Traveling-Salesperson-Problem (TSP) is an intriguing fundamental and well-studied $\mathcal{NP}$-hard optimization problem. Given a complete graph the TSP asks for a Hamiltonian cycle of minimum length, i.e., a round-trip salesperson tour that visits each node exactly once before ending at the start node. In the Euclidean TSP (E-TSP) nodes are associated with point coordinates in the Euclidean space and pairwise (symmetric) inter-node distances are given by the Euclidean distance; the E-TSP remains $\mathcal{NP}$-hard.


Since its introduction in 1930 a body of knowledge has been built around the TSP. As a consequence, a plethora of methods has been developed ranging from sophisticated exact solvers (guarantee to find an optimum) to fast heuristic algorithmic approaches with no performance guarantees at all. In the domain of exact TSP solving, the branch-and-cut based Concorde solver by \cite{applegate2007} is the state of the art. However, even though instances with hundreds of nodes can be solved within seconds~\citep{cook2012}, no guarantees for reasonable runtime can be given for large instances. 

For the E-TSP, distances adhere to the triangle inequality induced by the Euclidean metric. This property can be leveraged to come up with approximation algorithms. For a minimization problem, an $(1+\alpha)$-approximation algorithm $A$ guarantees that the tour length of $A$ on a problem instance is at most $(1 + \alpha) \cdot OPT$, where $OPT$ is the optimal tour length.
Christofides~\cite{christofides1976} introduced an algorithm that achieves an approximation factor of $3/2$, which is the best constant approximation factor known for the E-TSP. 

\newpage
Celebrated work by Arora~\cite{Arora1998} and -- independently -- by Mitchell~\citet{Mitchell1999} introduced a \emph{Polynomial Time Approximation Schema} (PTAS) algorithm for the metric TSP which guarantees to produce solutions of quality at most $(1 + 1/\alpha) \cdot OPT$ for each constant $\alpha > 0$ in polynomial time. However, the algorithms are highly sophisticated and to the best of our knowledge no practical implementation of this PTAS is available. Moreover, PTAS naturally suffers from impractical runtimes if $\alpha$ is increased -- in other words: a reduction of $\alpha$ goes hand in hand with a dramatic increase of the polynomial degree.

In recent years tremendous advances in heuristic TSP solving have been made where no formal performance guarantee can be given. Nevertheless the two best performing heuristics, LKH by \cite{helsgaun_general_2009} (based on sophisticated $k$-opt moves and the Lin-Kernighan heuristic) and the genetic algorithm EAX by \cite{Nagata2013} (a $(\mu + \lambda)$ genetic algorithm adopting the eponymous edge assembly crossover operator and sophisticated diversity preservation) solve instances with thousands of nodes to optimality within reasonable time limits~\citep{Hutter2015}. The respective restart versions LKH+r and EAX+r, which trigger a restart once the internal stopping conditions of the respective vanilla versions have been satisfied, pose the state of the art in inexact TSP solving \cite{KKHT2015, KKBHTLeveragingTSP}. Recent endeavours by \cite{Sanches2017} extend both solvers by a sophisticated crossover operator -- the generalized partition crossover (GPX2) -- which has shown superior performance over the vanilla version of EAX on large instances with node numbers in the five-digit range. 

In the field of per-instance algorithm selection (AS) -- see, e.g., the survey of \cite{kerschke2019survey} for further details -- the goal is to build a model which automatically selects the best-performing algorithm from a portfolio of algorithms with complementary performances. In case of the E-TSP, the complementary behavior of EAX(+r) and LKH(+r) across a wide range of problem instances has been leveraged in several studies in recent years \cite{KKHT2015, KKBHTLeveragingTSP}. Both of these works focused on optimality of the found solutions, i.e., the runtime until a solver found a tour of optimal (i.e., minimal) length was measured, and runs which did not succeed (within the given time of one hour) were penalized.

Despite their success in solving E-TSP instances up to optimality, little is known about the empirical approximation qualities, i.e., the anytime behaviour, of LKH+r and EAX+r building on the concept of empirical runtime distributions \cite{hoos2004stochastic}. Our work will thus shed light on the relationship between runtimes and respective approximation qualities. This is conceptually similar to the commonly accepted benchmarking practice in single-objective continuous optimization on the Black-Box Optimization Benchmark (BBOB, \cite{hansen2016cocoplat}). It should be noted, however, that despite simultaneous consideration of runtime and solution quality, the herein considered analysis of the anytime behavior of a solver (w.r.t.~its solution quality) differs substantially from the multi-objective approach that was taken in \cite{BKT2019}. Our analysis is supported by an extensive study on a wide range of TSP instances from the literature with $500$, $1\,000$ and $2\,000$ nodes, respectively. More precisely, we pursue three research questions in this work:
\begin{enumerate}
    \item[\textbf{R1}] Given different $\alpha$-values and time limits $T$, what is the probability to calculate an $(1+\alpha)$-approximate solution for different variants of LKH+r and EAX+r on a large and diverse set of E-TSP instances in time $T$?
    \item[\textbf{R2}] Which approximation quality $(1+\alpha)$ can we expect given a time limit $T$?
    \item[\textbf{R3}] How can automated algorithm selection approaches make use of information on anytime behaviour of TSP solvers?
\end{enumerate}

These questions are addressed in the remainder of the paper which is  organized as follows. Section~\ref{sec:methodology} introduces the methodology underlying the experimental study, including problem instances and algorithms. Results are presented and analyzed in Section~\ref{sec:experiments}, and the paper concludes with a discussion of promising future research directions in Section~\ref{sec:conclusion}.



\section{Methodology}
\label{sec:methodology}
In this section we detail the setup of our experimental study. 

\subsection{Problem instances}
Several AS-studies revealed -- which is in line with intuition -- that characteristics of TSP problems, such as clustering properties or the depth of a minimum spanning tree, may have a strong impact on the running time until an optimal solution is found~\citep{KKBHTLeveragingTSP}. It is legitimate to assume that this will also be true for the $(1+\alpha)$-approximation case. However, feature impacts and relations may change. Thus, to investigate -- and ideally support -- our assumptions empirically, we conducted an experimental study across a wide range of different E-TSP instances. For better comparability of our results, our setup is aligned with previous studies~\citep{KKHT2015,KKBHTLeveragingTSP,McMenemy2019} and hence covers the following TSP sets:

\begin{compactitem}
\item[\textbf{rue}] Classical \underline{R}andom \underline{U}niform \underline{E}uclidean instances where point coordinates are spread uniformly at random in the bounded Euclidean space $[0, 10^6] \times [0, 10^6]$.
\item[\textbf{netgen}] In \cite{MGBTR2015EvaluationOrienteering} this type of strongly clustered instances of size $n$ was proposed. For a given number of clusters $n_c \in \{2, 3, 4, 5, 10\}$, respective cluster centers are placed well-spread in the Euclidean plane (in $[0, 10^6] \times [0, 10^6]$) by Latin-Hypercube-Sampling (LHS). Subsequently, $n/n_c$ cities are sampled around each cluster center assuring cluster segregation.
\item[\textbf{morphed}] A morphed instance originates from combining a rue with a netgen instance of equal size. First, an optimal weighted point matching is calculated between the point coordinates of both instances. Next, the matched points are used to calculate new points by convex combination of the coordinates of the matched points. This approach was introduced in \cite{Mersmann2013} and later improved in \cite{MGBTR2015EvaluationOrienteering}. 
\item[\textbf{tspgen}] Instances were generated by sequential application of mutation to an initial rue instance as proposed by \cite{bossek2019}. For each mutation, a random subset of points is selected and rearranged by means of \enquote{creative} mutation operators, e.g., a grid-mutation, which aligns a random subset of points in a grid structure. These operators are inspired by observations on real-world instances (e.g., from Very Large Scale Integration, VLSI) and meant to produce instances that are structurally heterogeneous.
\item[\textbf{evolved}] TSP instances evolved by means of an evolutionary algorithm which minimizes the ratio of Penalized-Average-Runtime (PAR, \cite{Bischl2016}) scores\footnote{The PAR score is the average running time of $m$ independent solver runs until an optimal solution was found. Runs which are not successful within time $T$ are penalized with $f\cdot T$ where $f$ is a penalty factor usually set to 10.} of solvers EAX+r and LKH+r producing instances that are easy for one and hard(er) for the competitor. The set of evolved instances considered within this work is taken from \cite{bossek2019}.
\end{compactitem}
For instance sets rue, netgen, morphed and tspgen we each consider 150 instances of size $n \in \{500, 1\,500, 2000\}$. Subsets of 30 instances thus contain $n_c \in \{2, 3, 4, 5, 10\}$ clusters for instances of type netgen. The evolved instances -- taken from \cite{bossek2019} -- are restricted to $n=500$ nodes\footnote{Those instances were generated by an evolutionary algorithm where a single fitness function evaluation requires (1) a call of the exact Concorde solver and (2) multiple runs of LKH+r and EAX+r respectively. This becomes computationally very expensive for $n \in \{1\,000, 2\,000\}$.}. There are each 100 instances which are easy for EAX+r and LKH+r respectively. Summing up, in total, our benchmark set constitutes $2\,000$ E-TSP instances. 
Note that we intentionally do not include instances from the well-known TSPLIB~\citep{Reinelt91tsplib} benchmark set. To make well-founded statements about the research questions addressed in this work we require a large and systematic set of instances from different classes of equal size. However, TSPLIB instances are very heterogeneous in both size and structure which does not allow for proper evaluation.

\subsection{Considered Algorithms}
In total our study considers six different solvers for the E-TSP. The first four are restart variants of LKH \citep{helsgaun_general_2009} while the latter two are restart variants of EAX \citep{Nagata2013}\footnote{Restart variants trigger a cold restart once the internal stopping conditions are hit. This modification to LKH and EAX was introduced in \cite{KKHT2015}.}. In particular these variants incorporate generalized partition crossover (GPX2) into the algorithms \citep{Sanches2017}.
\begin{compactitem}
\item[\textbf{LKH variants:}] The LKH algorithm is an iterated local search algorithm that heuristically generates $k$-opt moves. A powerful improvement of LKH was the introduction of multi-trial LKH, where several solutions originating from soft restarts of the Lin-Kernigham heuristic are recombined by a partition crossover operator named Iterative Partial Transcription (IPT). A recent proposal replaces IPT by the alternative crossover operator GPX2. Additionally, LKH v2.0.9 allows to use both IPT and GPX2 in sequence. Therefore, we consider the four restarts variants LKH+r (IPT), i.e., the vanilla version of LKH+r, LKH+r (GPX), LKH+r (IPT+GPX) and LKH+r (GPX+IPT).
\item[\textbf{EAX variants:}] 
EAX is a powerful genetic algorithm which uses the Edge Assembly Crossover (EAX) operator to combine two parents. The operator is designed to keep as many edges from the parents as possible and introduces only a few short edges to complete the tour. The EAX algorithm is a $(\mu + \lambda)$-strategy with a sophisticated diversity preservation technique based on edge entropy to prevent the algorithm from premature convergence. We use the restart version EAX+r and additionally consider a modified version where individuals created by EAX+r are further recombined by applying GPX2. It should be noted that our modification is more straight-forward than the different variants introduced in \cite{Sanches2017}.
\end{compactitem}

\subsection{Estimation of Probabilities}
Next, we describe the process of probability estimation. Given a value $\alpha \geq 0$, a time-limit $T$, a stochastic algorithm $A$, and an instance $I$ with optimal tour length $\text{OPT}$, we denote the probability to reach a solution of desired quality within the
time-limit $T$ as $p_{\alpha, T}^A(I) = P\left(A(I) \leq (1 + \alpha) \cdot \text{OPT}\right)$. Given $A$ is stochastic and the trajectories of $m$ independent runs of that algorithm on instance $I$ are available, the probability $p_{\alpha, T}^{A}(I)$ can be estimated by the relative number of runs that succeeded in finding an $(1+\alpha)$-approximation within $T$, i.e.,
\begin{align*}
\hat{p}_{\alpha, T}^{A}(I) = \frac{1}{m} \sum_{i=1}^{m} \mathds{1}\Bigl(A^{i}_{T}(I) \leq (1+\alpha) \cdot \text{OPT}\Bigr).
\end{align*}
Here, $A^{i}_{T}(I)$ is the incumbent solution of $A$ on $I$ in the $i$-th run after time $T$ and $\mathds{1}$ is the indicator function evaluating to $1$ if its argument is true.
Given a set $\mathcal{I}$ of instances, an estimator for the success probability $p_{\alpha, T}^{A}(\mathcal{I})$ on the set $\mathcal{I}$ is the average probability over all instances in $\mathcal{I}$, i.e.,
$$
\hat{p}_{\alpha, T}^{A}(\mathcal{I}) = \frac{1}{|\mathcal{I}|} \sum_{I \in \mathcal{I}} \hat{p}_{\alpha, T}^{A}(I).
$$

\section{Experiments}
\label{sec:experiments}

\subsection{Experimental Setup}
Each of the six algorithms was run on each instance $m=10$ times with different random seeds in order to account for stochasticity. Throughout those experiments, we used a cutoff time of $T=3\,600$ seconds (i.e., one hour). The algorithms log the incumbent, i.e., best-so-far, solution in every run.

As all results for the TSP sets netgen, morphed and tspgen were qualitatively similar, we combined the respective information into a single set called ``structured''. 

\subsection{Perspective 1: First hitting times}

In practical applications, often an upper bound on solver performance is desired in order to realistically assess the worst case scenario. Therefore, it is of interest how long it will take a solver \emph{at most} to find a $(1+\alpha)$-approximation of the true optimum. 
The chosen solvers were executed for a variety of approximation gaps $\alpha$ and for each combination with TSP set and instance sizes $n \in \{500, 1\,000, 2\,000\}$. Note that in this part of our study, we analyze the algorithms' performances across a wide range of approximation factors, $\alpha \in \{0.5, 0.1, 0.05, 0.01, \ldots, 5\cdot 10^{-5}, 10^{-5}\}$. Thereby, we get a comprehensive overview of the behavior of the different algorithms, and later on can zoom in on more relevant areas. We take a pessimistic perspective and estimate the first hitting time as the \underline{maximum time} needed by each algorithm to find a solution of the corresponding quality $(1+\alpha)$ for the first time across all instances and runs of each instance set and size respectively.

\begin{figure*}[t]
    \centering
    \includegraphics[width=\textwidth, trim=0pt 7pt 0pt 15pt, clip]{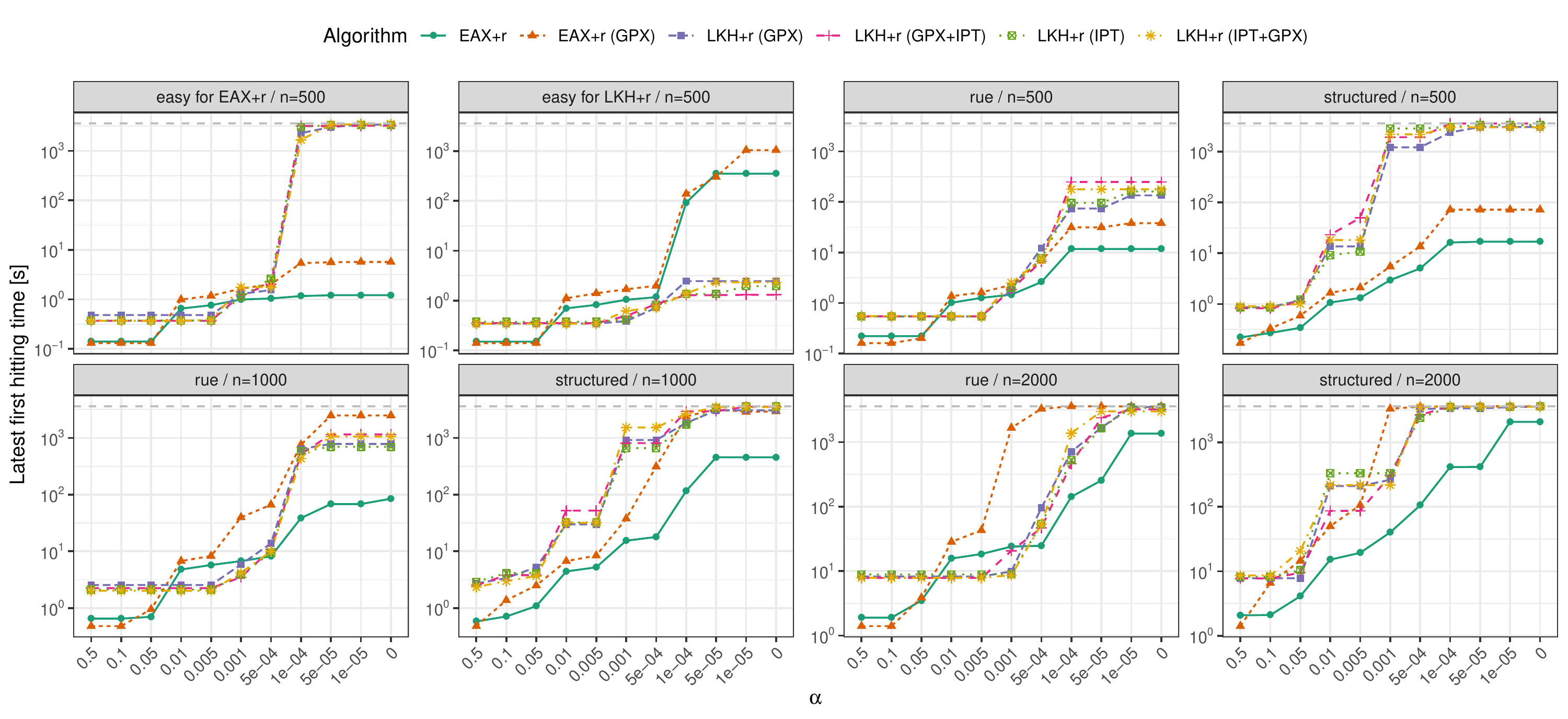}
    \caption{Maximum first hitting times for different $\alpha$-values. That is, we report for each approximation factor $\alpha$ ($x$-axis) the maximum time needed by EAX+r and LKH+r to find a solution of the corresponding quality for the first time across all instances and runs. It is therefore a maximally pessimistic view on approximation quality. Splits are done by instance class and size.}
    \label{fig:first_hitting_times}
\end{figure*}

As depicted in Fig.~\ref{fig:first_hitting_times}, both EAX variants are extremely fast for large values of $\alpha$, i.e., their solutions at early stages of the runs are already of very good quality. In fact, for $\alpha = 0.5$, EAX+r (GPX) had the lowest first hitting times across all eight considered scenarios, which supports the effectiveness of the sophisticated crossover operator GPX. More precisely, for small ($n = 500$) and medium-sized ($n = 1\,000$) instances, all runs of EAX+r (GPX) found a tour that is at most $50\%$ longer than the optimal tour within less than a second. On larger instances ($n = 2\,000$), this optimizer needed just slightly more than a second. Yet, it is also observable that the classical EAX+r performs better than its GPX-counterpart for decreasing $\alpha$-values and even outperforms it for all approximation gaps $\alpha \le 0.01$.

Further noticeable findings can be derived for LKH and its variants. First, the trajectories of all four LKH variants are almost identical across all the investigated scenarios, implying that no substantial differences among the considered versions regarding latest first hitting times could be detected. Moreover, with the exception of the medium-sized and large structured TSP instances with $n \in \{1\,000, 2\,000\}$ nodes, LKH performs at least as well as EAX within the mid-range approximation factors $\alpha \in [0.0005, 0.01]$. However, for the small approximation factors, LKH is again inferior to EAX -- except for the problems that were specifically tailored in favor of LKH. 


Thus, depending on the desired approximation quality, one could provide a three-fold recommendation: if the acceptable approximation-quality is rather large ($\alpha > 0.1$) EAX+r (ideally its GPX-enhanced version) is preferable, for mid-range values of $\alpha$ one should rather use one of the LKH variants, and for very small approximation factors ($\alpha \le 0.0001$) one should consider the classical EAX+r. However, it has to be kept in mind that these findings solely focus the worst case scenario. 



\subsection{Perspective 2: Probability of Success}

While the previous subsection focused on the latest first hitting time, i.e., a worst case analysis, we are now interested in the solvers' average case performances. Of course, previous studies already addressed the average case as well, but usually with a focus on (penalized) aggregations of runtimes. Unfortunately, those runtimes only considered whether an algorithm found a tour of optimal length -- which corresponds to $\alpha=0$. Further information on the solver's performance, such as whether it failed (to find an optimal tour) by orders of magnitude or just by an infinitesimal deviation was neglected. In the following, we will overcome this ``knowledge gap'' by comparing the performances of a total of six versions of EAX+r and LKH+r across different TSP sets and approximation factors. More precisely, within this subsection we investigate  the change of an algorithm's average success probability over time for different approximation gaps. For small instances ($n = 500$) we used $\alpha \in [0, 10^{-4}]$ and otherwise $\alpha \in [0, 10^{-3}]$. This choice of $\alpha$-values was considered sufficiently large, as for the largest values in those intervals the average success probability of most algorithms already converged to 1.0 within the investigated time.

\begin{figure*}[t]
    \centering
    \includegraphics[width=\textwidth, trim=0pt 0pt 0pt 11pt, clip]{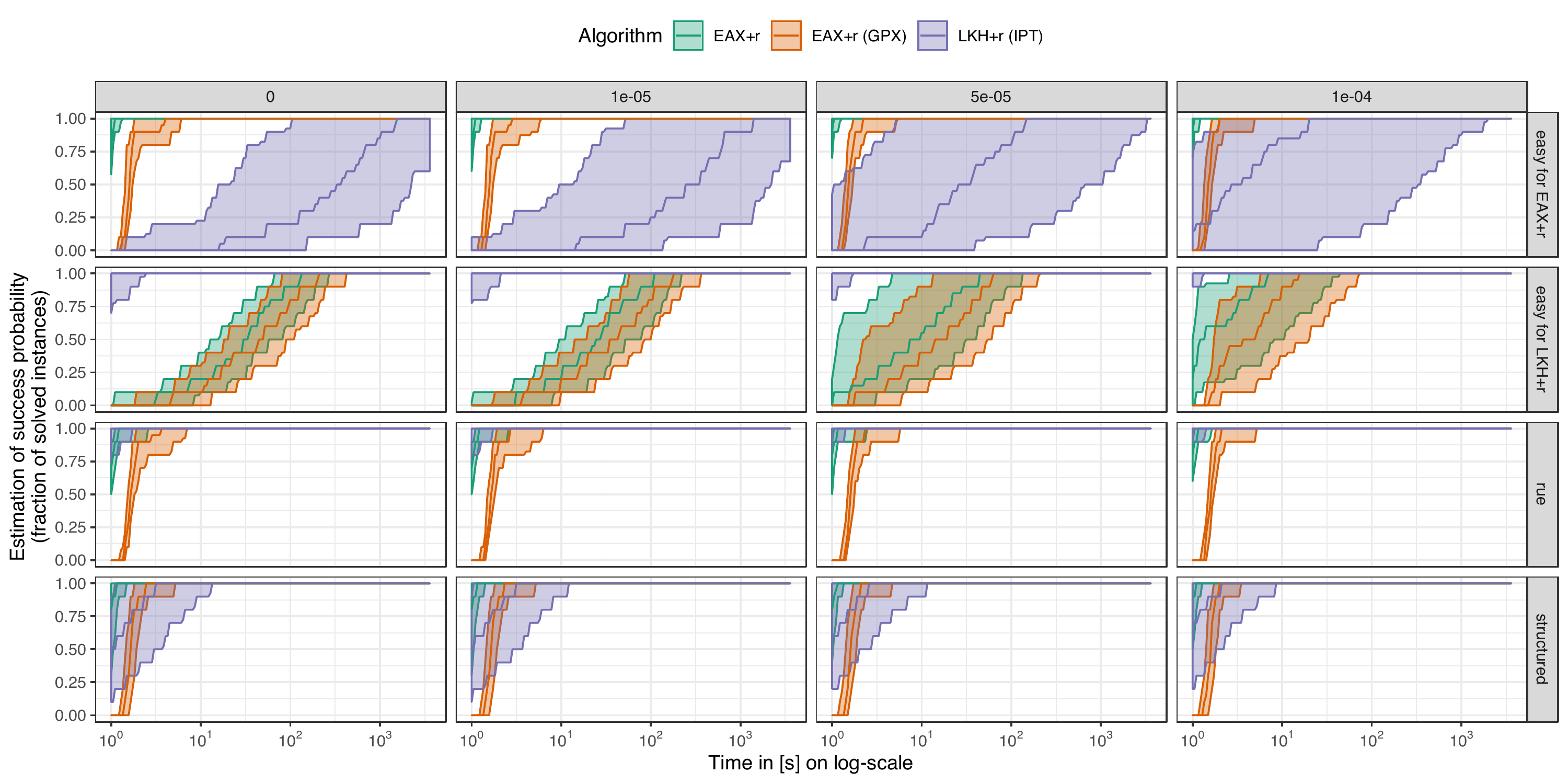}
    \caption{
    This plot shows the median success probabilities to locate a solution of quality $(1+\alpha)\cdot \text{OPT}$ (columns show different $\alpha$ values) for instances with $n = 500$ nodes. The tubes are defined by the corresponding $0.25$-quantile and $0.75$-quantile, respectively.} 
    \label{fig:lineplot_mean_sets_by_time_and_gap_n500}
\end{figure*}

The results for the TSP instances with $n = 500$ nodes are depicted in Fig.~\ref{fig:lineplot_mean_sets_by_time_and_gap_n500} and also listed in Tab.~\ref{tab:probs_small}. As the performances of all four LKH variants are very similar only the results for LKH+r's default version, i.e., LKH+r (IPT) are provided as a representative.
Interestingly, LKH+r outperforms both EAX variants on the TSP set ``easy for LKH+r'' across all considered combinations of time steps and approximation gaps, although those instances have been generated for just one specific pair of maximum runtime ($T = 5$ minutes) and approximation quality ($\alpha = 0$) as detailed in \cite{bossek2019}.
These findings are reflected by the median probability curves depicted in Fig.~\ref{fig:lineplot_mean_sets_by_time_and_gap_n500}. As shown in the second row of this plot, the curve of LKH+r is located in the very top-left corner, implying that it has achieved a high success probability (w.r.t. the respective approximation quality) within seconds. In contrast, the ``tubes'' associated with the EAX+r variants are rather broad and spread diagonally across all four images. That is, their median success probabilities show a high variance and both solvers require much more time to achieve feasible success probabilities (closer to 1.00) on those instances. The superiority of LKH+r over the two EAX+r versions is also confirmed by pairwise Wilcoxon-tests to a significance level of $5\%$ (as indicated by the \textcolor{colorEAXR}{$\text{1}^{+}$} and \textcolor{colorEAXRGPX}{$\text{2}^{+}$} in the last column of Tab.~\ref{tab:probs_small}).

On the other hand, the two EAX versions are absolutely superior to LKH+r on the ``easy for EAX+r'' problems. In fact, the tube of LKH+r basically covers the majority of each of the images in the first row of Fig.~\ref{fig:lineplot_mean_sets_by_time_and_gap_n500}. 

We further noticed that while EAX+r and its variant performs well on all sets except for ``easy for LKH'', LKH+r exhibits clear preferences -- ranging from very poor performances on the EAX-tailored instances, via mediocre behavior on the structured instances, up to good performances on rue, and (of course) the LKH-tailored problems. This clearly indicates that the structural properties of an instance, i.e., its node alignment, strongly affect the optimization behavior of LKH.




\begin{scriptsize}
\begin{table*}[htbp]

\renewcommand{\arraystretch}{1.1}
\renewcommand{\tabcolsep}{6pt}
\caption{\label{tab:probs_small}Maximum gap (\textbf{max}), mean success probability (\textbf{mean}), standard deviation (\textbf{std}) and results of pairwise Wilcoxon-tests for EAX+r, EAX+r (GPX) and LKH+r (IPT). A value $X^+$ in the stat column indicates that the results of the algorithm are statistically significant in comparison to algorithm $X$. Results are split by instance group, $\alpha$ and time for instances of size \underline{$n=500$}. Best mean values per row are highlighted in \colorbox{gray!20}{\textbf{bold face}}.}
\centering
\begin{scriptsize}
\begin{tabular}[t]{lrrrrrrrrrrrrrrr}
\toprule
\multicolumn{1}{c}{\textbf{ }} & \multicolumn{1}{c}{\textbf{ }} & \multicolumn{1}{c}{\textbf{ }} & \multicolumn{1}{c}{\textbf{ }} & \multicolumn{4}{c}{\textcolor{colorEAXR}{\textbf{EAX+r (1)}}} & \multicolumn{4}{c}{\textcolor{colorEAXRGPX}{\textbf{EAX+r (GPX) (2)}}} & \multicolumn{4}{c}{\textcolor{colorLKHRIPT}{\textbf{LKH+r (IPT) (3)}}} \\
\cmidrule(l{3pt}r{3pt}){5-8} \cmidrule(l{3pt}r{3pt}){9-12} \cmidrule(l{3pt}r{3pt}){13-16}
\textbf{Group} & $n$ & $\alpha$ & $T$ & \textbf{max} & \textbf{mean} & \textbf{std} & \textbf{stat} & \textbf{max} & \textbf{mean} & \textbf{std} & \textbf{stat} & \textbf{max} & \textbf{mean} & \textbf{std} & \textbf{stat}\\
\midrule
 & 500 & $0.00010$ & 10 & $0.00$ & \cellcolor{gray!20}{\textbf{1.00}} & 0.00 & \textcolor{colorEAXRGPX}{$\text{2}^{+}$}, \textcolor{colorLKHRIPT}{$\text{3}^{+}$} & 0 & 0.99 & 0.04 & \textcolor{colorLKHRIPT}{$\text{3}^{+}$} & 0.00 & 0.55 & 0.46 & \\

 & 500 & $0.00010$ & 50 & $0.00$ & \cellcolor{gray!20}{\textbf{1.00}} & 0.00 & \textcolor{colorLKHRIPT}{$\text{3}^{+}$} & 0 & \cellcolor{gray!20}{\textbf{1.00}} & 0.00 & \textcolor{colorLKHRIPT}{$\text{3}^{+}$} & 0.00 & 0.66 & 0.44 & \\

 & 500 & $0.00010$ & 100 & $0.00$ & \cellcolor{gray!20}{\textbf{1.00}} & 0.00 & \textcolor{colorLKHRIPT}{$\text{3}^{+}$} & 0 & \cellcolor{gray!20}{\textbf{1.00}} & 0.00 & \textcolor{colorLKHRIPT}{$\text{3}^{+}$} & 0.00 & 0.69 & 0.42 & \\

\cmidrule{3-16}
 & 500 & $0.00005$ & 10 & $0.00$ & \cellcolor{gray!20}{\textbf{1.00}} & 0.00 & \textcolor{colorEAXRGPX}{$\text{2}^{+}$}, \textcolor{colorLKHRIPT}{$\text{3}^{+}$} & 0 & 0.99 & 0.04 & \textcolor{colorLKHRIPT}{$\text{3}^{+}$} & 0.00 & 0.42 & 0.46 & \\

 & 500 & $0.00005$ & 50 & $0.00$ & \cellcolor{gray!20}{\textbf{1.00}} & 0.00 & \textcolor{colorLKHRIPT}{$\text{3}^{+}$} & 0 & \cellcolor{gray!20}{\textbf{1.00}} & 0.00 & \textcolor{colorLKHRIPT}{$\text{3}^{+}$} & 0.00 & 0.54 & 0.45 & \\

 & 500 & $0.00005$ & 100 & $0.00$ & \cellcolor{gray!20}{\textbf{1.00}} & 0.00 & \textcolor{colorLKHRIPT}{$\text{3}^{+}$} & 0 & \cellcolor{gray!20}{\textbf{1.00}} & 0.00 & \textcolor{colorLKHRIPT}{$\text{3}^{+}$} & 0.00 & 0.59 & 0.44 & \\

\cmidrule{3-16}
 & 500 & $0.00001$ & 10 & $0.00$ & \cellcolor{gray!20}{\textbf{1.00}} & 0.00 & \textcolor{colorEAXRGPX}{$\text{2}^{+}$}, \textcolor{colorLKHRIPT}{$\text{3}^{+}$} & 0 & 0.99 & 0.06 & \textcolor{colorLKHRIPT}{$\text{3}^{+}$} & 0.00 & 0.27 & 0.40 & \\

 & 500 & $0.00001$ & 50 & $0.00$ & \cellcolor{gray!20}{\textbf{1.00}} & 0.00 & \textcolor{colorLKHRIPT}{$\text{3}^{+}$} & 0 & 1.00 & 0.01 & \textcolor{colorLKHRIPT}{$\text{3}^{+}$} & 0.00 & 0.38 & 0.43 & \\

 & 500 & $0.00001$ & 100 & $0.00$ & \cellcolor{gray!20}{\textbf{1.00}} & 0.00 & \textcolor{colorLKHRIPT}{$\text{3}^{+}$} & 0 & \cellcolor{gray!20}{\textbf{1.00}} & 0.00 & \textcolor{colorLKHRIPT}{$\text{3}^{+}$} & 0.00 & 0.43 & 0.43 & \\

\cmidrule{3-16}
 & 500 & $0.00000$ & 10 & $0.00$ & \cellcolor{gray!20}{\textbf{1.00}} & 0.00 & \textcolor{colorEAXRGPX}{$\text{2}^{+}$}, \textcolor{colorLKHRIPT}{$\text{3}^{+}$} & 0 & 0.99 & 0.06 & \textcolor{colorLKHRIPT}{$\text{3}^{+}$} & 0.00 & 0.24 & 0.38 & \\

 & 500 & $0.00000$ & 50 & $0.00$ & \cellcolor{gray!20}{\textbf{1.00}} & 0.00 & \textcolor{colorLKHRIPT}{$\text{3}^{+}$} & 0 & 1.00 & 0.01 & \textcolor{colorLKHRIPT}{$\text{3}^{+}$} & 0.00 & 0.35 & 0.41 & \\

\multirow{-20}{*}{\raggedright\arraybackslash \textbf{easy for EAX+r}} & 500 & $0.00000$ & 100 & $0.00$ & \cellcolor{gray!20}{\textbf{1.00}} & 0.00 & \textcolor{colorLKHRIPT}{$\text{3}^{+}$} & 0 & \cellcolor{gray!20}{\textbf{1.00}} & 0.00 & \textcolor{colorLKHRIPT}{$\text{3}^{+}$} & 0.00 & 0.40 & 0.41 & \\
\cmidrule{1-16}
 & 500 & $0.00010$ & 10 & $0.00$ & 0.77 & 0.34 & \textcolor{colorEAXRGPX}{$\text{2}^{+}$} & 0 & 0.68 & 0.37 &  & 0.00 & \cellcolor{gray!20}{\textbf{1.00}} & 0.01 & \textcolor{colorEAXR}{$\text{1}^{+}$}, \textcolor{colorEAXRGPX}{$\text{2}^{+}$}\\

 & 500 & $0.00010$ & 50 & $0.00$ & 0.90 & 0.20 &  & 0 & 0.85 & 0.28 &  & 0.00 & \cellcolor{gray!20}{\textbf{1.00}} & 0.00 & \textcolor{colorEAXR}{$\text{1}^{+}$}, \textcolor{colorEAXRGPX}{$\text{2}^{+}$}\\

 & 500 & $0.00010$ & 100 & $0.00$ & 0.95 & 0.14 &  & 0 & 0.92 & 0.18 &  & 0.00 & \cellcolor{gray!20}{\textbf{1.00}} & 0.00 & \textcolor{colorEAXR}{$\text{1}^{+}$}, \textcolor{colorEAXRGPX}{$\text{2}^{+}$}\\

\cmidrule{3-16}
 & 500 & $0.00005$ & 10 & $0.00$ & 0.57 & 0.39 & \textcolor{colorEAXRGPX}{$\text{2}^{+}$} & 0 & 0.48 & 0.38 &  & 0.00 & \cellcolor{gray!20}{\textbf{1.00}} & 0.01 & \textcolor{colorEAXR}{$\text{1}^{+}$}, \textcolor{colorEAXRGPX}{$\text{2}^{+}$}\\

 & 500 & $0.00005$ & 50 & $0.00$ & 0.81 & 0.27 & \textcolor{colorEAXRGPX}{$\text{2}^{+}$} & 0 & 0.73 & 0.31 &  & 0.00 & \cellcolor{gray!20}{\textbf{1.00}} & 0.00 & \textcolor{colorEAXR}{$\text{1}^{+}$}, \textcolor{colorEAXRGPX}{$\text{2}^{+}$}\\

 & 500 & $0.00005$ & 100 & $0.00$ & 0.89 & 0.21 &  & 0 & 0.86 & 0.22 &  & 0.00 & \cellcolor{gray!20}{\textbf{1.00}} & 0.00 & \textcolor{colorEAXR}{$\text{1}^{+}$}, \textcolor{colorEAXRGPX}{$\text{2}^{+}$}\\

\cmidrule{3-16}
 & 500 & $0.00001$ & 10 & $0.00$ & 0.32 & 0.31 &  & 0 & 0.26 & 0.27 &  & 0.00 & \cellcolor{gray!20}{\textbf{1.00}} & 0.01 & \textcolor{colorEAXR}{$\text{1}^{+}$}, \textcolor{colorEAXRGPX}{$\text{2}^{+}$}\\

 & 500 & $0.00001$ & 50 & $0.00$ & 0.67 & 0.32 & \textcolor{colorEAXRGPX}{$\text{2}^{+}$} & 0 & 0.57 & 0.32 &  & 0.00 & \cellcolor{gray!20}{\textbf{1.00}} & 0.00 & \textcolor{colorEAXR}{$\text{1}^{+}$}, \textcolor{colorEAXRGPX}{$\text{2}^{+}$}\\

 & 500 & $0.00001$ & 100 & $0.00$ & 0.79 & 0.27 & \textcolor{colorEAXRGPX}{$\text{2}^{+}$} & 0 & 0.73 & 0.29 &  & 0.00 & \cellcolor{gray!20}{\textbf{1.00}} & 0.00 & \textcolor{colorEAXR}{$\text{1}^{+}$}, \textcolor{colorEAXRGPX}{$\text{2}^{+}$}\\

\cmidrule{3-16}
 & 500 & $0.00000$ & 10 & $0.00$ & 0.26 & 0.24 &  & 0 & 0.21 & 0.22 &  & 0.00 & \cellcolor{gray!20}{\textbf{1.00}} & 0.01 & \textcolor{colorEAXR}{$\text{1}^{+}$}, \textcolor{colorEAXRGPX}{$\text{2}^{+}$}\\

 & 500 & $0.00000$ & 50 & $0.00$ & 0.64 & 0.31 & \textcolor{colorEAXRGPX}{$\text{2}^{+}$} & 0 & 0.53 & 0.32 &  & 0.00 & \cellcolor{gray!20}{\textbf{1.00}} & 0.00 & \textcolor{colorEAXR}{$\text{1}^{+}$}, \textcolor{colorEAXRGPX}{$\text{2}^{+}$}\\

\multirow{-20}{*}{\raggedright\arraybackslash \textbf{easy for LKH+r}} & 500 & $0.00000$ & 100 & $0.00$ & 0.78 & 0.27 & \textcolor{colorEAXRGPX}{$\text{2}^{+}$} & 0 & 0.69 & 0.31 &  & 0.00 & \cellcolor{gray!20}{\textbf{1.00}} & 0.00 & \textcolor{colorEAXR}{$\text{1}^{+}$}, \textcolor{colorEAXRGPX}{$\text{2}^{+}$}\\
\cmidrule{1-16}
 & 500 & $0.00010$ & 10 & $0.00$ & \cellcolor{gray!20}{\textbf{1.00}} & 0.01 & \textcolor{colorEAXRGPX}{$\text{2}^{+}$}, \textcolor{colorLKHRIPT}{$\text{3}^{+}$} & 0 & 0.99 & 0.04 &  & 0.00 & 0.99 & 0.07 & \\

 & 500 & $0.00010$ & 50 & $0.00$ & \cellcolor{gray!20}{\textbf{1.00}} & 0.00 &  & 0 & \cellcolor{gray!20}{\textbf{1.00}} & 0.00 &  & 0.00 & 1.00 & 0.03 & \\

 & 500 & $0.00010$ & 100 & $0.00$ & \cellcolor{gray!20}{\textbf{1.00}} & 0.00 &  & 0 & \cellcolor{gray!20}{\textbf{1.00}} & 0.00 &  & 0.00 & \cellcolor{gray!20}{\textbf{1.00}} & 0.00 & \\

\cmidrule{3-16}
 & 500 & $0.00005$ & 10 & $0.00$ & \cellcolor{gray!20}{\textbf{1.00}} & 0.01 & \textcolor{colorEAXRGPX}{$\text{2}^{+}$}, \textcolor{colorLKHRIPT}{$\text{3}^{+}$} & 0 & 0.99 & 0.05 &  & 0.00 & 0.99 & 0.07 & \\

 & 500 & $0.00005$ & 50 & $0.00$ & \cellcolor{gray!20}{\textbf{1.00}} & 0.00 &  & 0 & \cellcolor{gray!20}{\textbf{1.00}} & 0.00 &  & 0.00 & 1.00 & 0.03 & \\

 & 500 & $0.00005$ & 100 & $0.00$ & \cellcolor{gray!20}{\textbf{1.00}} & 0.00 &  & 0 & \cellcolor{gray!20}{\textbf{1.00}} & 0.00 &  & 0.00 & \cellcolor{gray!20}{\textbf{1.00}} & 0.00 & \\

\cmidrule{3-16}
 & 500 & $0.00001$ & 10 & $0.00$ & \cellcolor{gray!20}{\textbf{1.00}} & 0.01 & \textcolor{colorEAXRGPX}{$\text{2}^{+}$}, \textcolor{colorLKHRIPT}{$\text{3}^{+}$} & 0 & 0.99 & 0.05 &  & 0.00 & 0.98 & 0.09 & \\

 & 500 & $0.00001$ & 50 & $0.00$ & \cellcolor{gray!20}{\textbf{1.00}} & 0.00 &  & 0 & \cellcolor{gray!20}{\textbf{1.00}} & 0.00 &  & 0.00 & 1.00 & 0.04 & \\

 & 500 & $0.00001$ & 100 & $0.00$ & \cellcolor{gray!20}{\textbf{1.00}} & 0.00 &  & 0 & \cellcolor{gray!20}{\textbf{1.00}} & 0.00 &  & 0.00 & 1.00 & 0.01 & \\

\cmidrule{3-16}
 & 500 & $0.00000$ & 10 & $0.00$ & \cellcolor{gray!20}{\textbf{1.00}} & 0.01 & \textcolor{colorEAXRGPX}{$\text{2}^{+}$}, \textcolor{colorLKHRIPT}{$\text{3}^{+}$} & 0 & 0.98 & 0.06 &  & 0.00 & 0.98 & 0.09 & \\

 & 500 & $0.00000$ & 50 & $0.00$ & \cellcolor{gray!20}{\textbf{1.00}} & 0.00 &  & 0 & \cellcolor{gray!20}{\textbf{1.00}} & 0.00 &  & 0.00 & 1.00 & 0.04 & \\

\multirow{-20}{*}{\raggedright\arraybackslash \textbf{rue}} & 500 & $0.00000$ & 100 & $0.00$ & \cellcolor{gray!20}{\textbf{1.00}} & 0.00 &  & 0 & \cellcolor{gray!20}{\textbf{1.00}} & 0.00 &  & 0.00 & 1.00 & 0.01 & \\
\cmidrule{1-16}
 & 500 & $0.00010$ & 10 & $0.00$ & \cellcolor{gray!20}{\textbf{1.00}} & 0.01 & \textcolor{colorEAXRGPX}{$\text{2}^{+}$}, \textcolor{colorLKHRIPT}{$\text{3}^{+}$} & 0 & 1.00 & 0.03 & \textcolor{colorLKHRIPT}{$\text{3}^{+}$} & 0.01 & 0.91 & 0.22 & \\

 & 500 & $0.00010$ & 50 & $0.00$ & \cellcolor{gray!20}{\textbf{1.00}} & 0.00 & \textcolor{colorLKHRIPT}{$\text{3}^{+}$} & 0 & 1.00 & 0.00 & \textcolor{colorLKHRIPT}{$\text{3}^{+}$} & 0.00 & 0.97 & 0.13 & \\

 & 500 & $0.00010$ & 100 & $0.00$ & \cellcolor{gray!20}{\textbf{1.00}} & 0.00 & \textcolor{colorLKHRIPT}{$\text{3}^{+}$} & 0 & \cellcolor{gray!20}{\textbf{1.00}} & 0.00 & \textcolor{colorLKHRIPT}{$\text{3}^{+}$} & 0.00 & 0.98 & 0.11 & \\

\cmidrule{3-16}
 & 500 & $0.00005$ & 10 & $0.00$ & \cellcolor{gray!20}{\textbf{1.00}} & 0.01 & \textcolor{colorEAXRGPX}{$\text{2}^{+}$}, \textcolor{colorLKHRIPT}{$\text{3}^{+}$} & 0 & 0.99 & 0.04 & \textcolor{colorLKHRIPT}{$\text{3}^{+}$} & 0.01 & 0.89 & 0.24 & \\

 & 500 & $0.00005$ & 50 & $0.00$ & \cellcolor{gray!20}{\textbf{1.00}} & 0.00 & \textcolor{colorLKHRIPT}{$\text{3}^{+}$} & 0 & 1.00 & 0.00 & \textcolor{colorLKHRIPT}{$\text{3}^{+}$} & 0.00 & 0.96 & 0.15 & \\

 & 500 & $0.00005$ & 100 & $0.00$ & \cellcolor{gray!20}{\textbf{1.00}} & 0.00 & \textcolor{colorLKHRIPT}{$\text{3}^{+}$} & 0 & \cellcolor{gray!20}{\textbf{1.00}} & 0.00 & \textcolor{colorLKHRIPT}{$\text{3}^{+}$} & 0.00 & 0.98 & 0.13 & \\

\cmidrule{3-16}
 & 500 & $0.00001$ & 10 & $0.00$ & \cellcolor{gray!20}{\textbf{1.00}} & 0.01 & \textcolor{colorEAXRGPX}{$\text{2}^{+}$}, \textcolor{colorLKHRIPT}{$\text{3}^{+}$} & 0 & 0.99 & 0.04 & \textcolor{colorLKHRIPT}{$\text{3}^{+}$} & 0.01 & 0.87 & 0.26 & \\

 & 500 & $0.00001$ & 50 & $0.00$ & \cellcolor{gray!20}{\textbf{1.00}} & 0.00 & \textcolor{colorLKHRIPT}{$\text{3}^{+}$} & 0 & 1.00 & 0.00 & \textcolor{colorLKHRIPT}{$\text{3}^{+}$} & 0.00 & 0.96 & 0.16 & \\

 & 500 & $0.00001$ & 100 & $0.00$ & \cellcolor{gray!20}{\textbf{1.00}} & 0.00 & \textcolor{colorLKHRIPT}{$\text{3}^{+}$} & 0 & \cellcolor{gray!20}{\textbf{1.00}} & 0.00 & \textcolor{colorLKHRIPT}{$\text{3}^{+}$} & 0.00 & 0.97 & 0.13 & \\

\cmidrule{3-16}
 & 500 & $0.00000$ & 10 & $0.00$ & \cellcolor{gray!20}{\textbf{1.00}} & 0.01 & \textcolor{colorEAXRGPX}{$\text{2}^{+}$}, \textcolor{colorLKHRIPT}{$\text{3}^{+}$} & 0 & 0.99 & 0.04 & \textcolor{colorLKHRIPT}{$\text{3}^{+}$} & 0.01 & 0.87 & 0.26 & \\

 & 500 & $0.00000$ & 50 & $0.00$ & \cellcolor{gray!20}{\textbf{1.00}} & 0.00 & \textcolor{colorLKHRIPT}{$\text{3}^{+}$} & 0 & 1.00 & 0.00 & \textcolor{colorLKHRIPT}{$\text{3}^{+}$} & 0.00 & 0.96 & 0.16 & \\

\multirow{-20}{*}{\raggedright\arraybackslash \textbf{structured}} & 500 & $0.00000$ & 100 & $0.00$ & \cellcolor{gray!20}{\textbf{1.00}} & 0.00 & \textcolor{colorLKHRIPT}{$\text{3}^{+}$} & 0 & \cellcolor{gray!20}{\textbf{1.00}} & 0.00 & \textcolor{colorLKHRIPT}{$\text{3}^{+}$} & 0.00 & 0.97 & 0.13 & \\
\bottomrule
\end{tabular}
\end{scriptsize}
\end{table*}
\end{scriptsize}

\begin{figure*}[t]
    \centering
    \includegraphics[width=\textwidth, trim=0pt 30pt 0pt 11pt, clip]{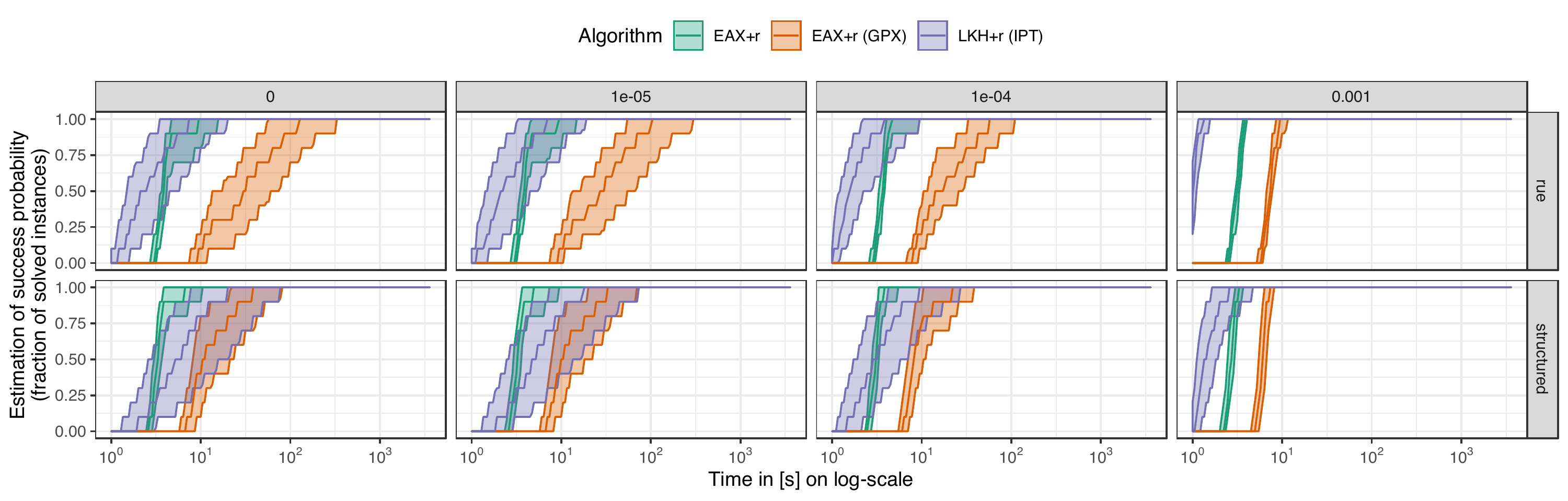}
    \includegraphics[width=\textwidth, trim=0pt 5pt 0pt 29pt, clip]{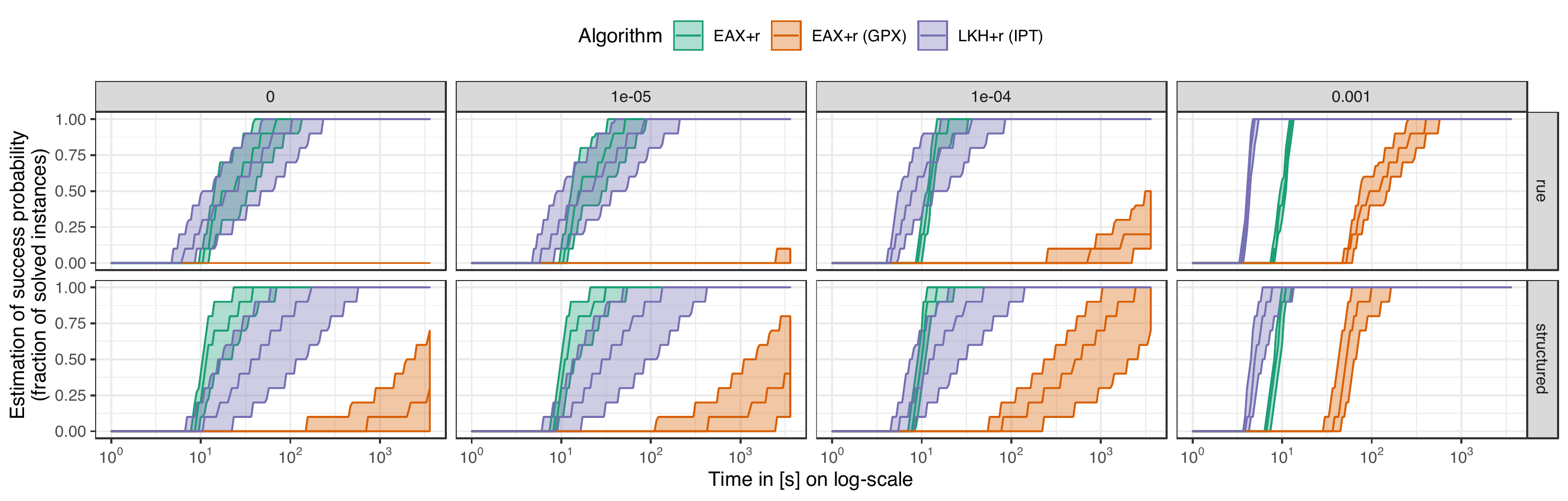}
    \caption{This plot shows the median success probabilities to locate a solution of quality $(1+\alpha)\cdot \text{OPT}$ (columns show different $\alpha$ values) for instances with $n = 1\,000$ nodes (two top rows) and $n =2\,000$ nodes (two bottom rows). The tubes are defined by the $0.25$-quantile and $0.75$-quantile, respectively.}
    \label{fig:lineplot_mean_sets_by_time_and_gap_g1000}
\end{figure*}

In contrast to our a priori expectations, we could not detect strong differences of a solver's success probabilities across the different approximation qualities for the small instances ($n = 500$) at hand. However, for  larger instances with $n \in \{1\,000, 2\,000\}$ nodes the picture slightly changes as shown in Fig.~\ref{fig:lineplot_mean_sets_by_time_and_gap_g1000} and Tab.~\ref{tab:probs_large}. First, with increasing instance size $n$ and decreasing $\alpha$, the EAX-variant EAX+r (GPX) is more and more outperformed by its contenders. This effect is reflected by the large number of $\textcolor{colorEAXR}{2^{+}}$'s in Tab.~\ref{tab:probs_large}, which indicate a significantly better performance of the respective solver compared to EAX+r (GPX). Secondly, across all solvers one can observe an increase in variation of the success probabilities (i.e., wider tubes), as well as slightly increasing runtimes (i.e., shift to the right) for decreasing approximation factors $\alpha$. However, those findings are quite intuitive as diminishing gaps correspond to more accurate results -- which are harder to find.

Observing Tab.~\ref{tab:probs_large} another pattern becomes visible: EAX+r performs significantly better than LKH+r on the structured instances. An even more interesting pattern can be observed for the (unstructured) rue instances. If the accepted approximation factor is rather small ($\alpha \le 0.0001$) and a sufficiently large time ($T \ge 50$s) is given, then EAX+r performs significantly better than LKH+r. However, if the instance size is large ($n = 2\,000$) and only a short amount of time is given ($T = 10$s), LKH+r is superior to both (considered) versions of EAX.

As a result, the ideal choice of (inexact) TSP optimization heuristic depends on a combination of (i) the magnitude of the allowed approximation gap, (ii) the size of the TSP instance, (iii) the given cutoff time for the solver, as well as (iv) the structure (i.e., node placement) of the instance itself. With the exception of (i), these findings are in line with the automated algorithm selection studies by \cite{KKHT2015,KKBHTLeveragingTSP}, in which the authors showed (for $\alpha = 0$) that an instance's structural information can be efficiently exploited to select the best performing optimization algorithm for the instance at hand.

\begin{scriptsize}
\begin{table*}[htbp]

\renewcommand{\arraystretch}{1.1}
\renewcommand{\tabcolsep}{6pt}
\caption{\label{tab:probs_large}Maximum gap (\textbf{max}), mean success probability (\textbf{mean}), standard deviation (\textbf{std}) and results of pairwise Wilcoxon-tests for EAX+r, EAX+r (GPX) and LKH+r (IPT). A value $X^+$ in the stat column indicates that the results of the algorithm are statistically significant in comparison to algorithm $X$. Results are split by instance group, $\alpha$ and time for instances of size \underline{$n>500$}. Best mean values per row are highlighted in \colorbox{gray!20}{\textbf{bold face}}.}
\centering
\begin{scriptsize}
\begin{tabular}[t]{lrrrrrrrrrrrrrrr}
\toprule
\multicolumn{1}{c}{\textbf{ }} & \multicolumn{1}{c}{\textbf{ }} & \multicolumn{1}{c}{\textbf{ }} & \multicolumn{1}{c}{\textbf{ }} & \multicolumn{4}{c}{\textcolor{colorEAXR}{\textbf{EAX+r (1)}}} & \multicolumn{4}{c}{\textcolor{colorEAXRGPX}{\textbf{EAX+r (GPX) (2)}}} & \multicolumn{4}{c}{\textcolor{colorLKHRIPT}{\textbf{LKH+r (IPT) (3)}}} \\
\cmidrule(l{3pt}r{3pt}){5-8} \cmidrule(l{3pt}r{3pt}){9-12} \cmidrule(l{3pt}r{3pt}){13-16}
\textbf{Group} & $n$ & $\alpha$ & $T$ & \textbf{max} & \textbf{mean} & \textbf{std} & \textbf{stat} & \textbf{max} & \textbf{mean} & \textbf{std} & \textbf{stat} & \textbf{max} & \textbf{mean} & \textbf{std} & \textbf{stat}\\
\midrule
 & 1000 & $0.00100$ & 10 & $0.00$ & \cellcolor{gray!20}{\textbf{1.00}} & 0.00 & \textcolor{colorEAXRGPX}{$\text{2}^{+}$} & 0.00 & 0.92 & 0.13 &  & 0.00 & \cellcolor{gray!20}{\textbf{1.00}} & 0.00 & \textcolor{colorEAXRGPX}{$\text{2}^{+}$}\\

 & 1000 & $0.00100$ & 50 & $0.00$ & \cellcolor{gray!20}{\textbf{1.00}} & 0.00 &  & 0.00 & \cellcolor{gray!20}{\textbf{1.00}} & 0.00 &  & 0.00 & \cellcolor{gray!20}{\textbf{1.00}} & 0.00 & \\

 & 1000 & $0.00100$ & 100 & $0.00$ & \cellcolor{gray!20}{\textbf{1.00}} & 0.00 &  & 0.00 & \cellcolor{gray!20}{\textbf{1.00}} & 0.00 &  & 0.00 & \cellcolor{gray!20}{\textbf{1.00}} & 0.00 & \\

\cmidrule{3-16}
 & 1000 & $0.00010$ & 10 & $0.00$ & \cellcolor{gray!20}{\textbf{0.97}} & 0.08 & \textcolor{colorEAXRGPX}{$\text{2}^{+}$} & 0.00 & 0.27 & 0.24 &  & 0.00 & 0.93 & 0.18 & \textcolor{colorEAXRGPX}{$\text{2}^{+}$}\\

 & 1000 & $0.00010$ & 50 & $0.00$ & \cellcolor{gray!20}{\textbf{1.00}} & 0.00 & \textcolor{colorEAXRGPX}{$\text{2}^{+}$}, \textcolor{colorLKHRIPT}{$\text{3}^{+}$} & 0.00 & 0.82 & 0.24 &  & 0.00 & 0.98 & 0.09 & \textcolor{colorEAXRGPX}{$\text{2}^{+}$}\\

 & 1000 & $0.00010$ & 100 & $0.00$ & \cellcolor{gray!20}{\textbf{1.00}} & 0.00 & \textcolor{colorEAXRGPX}{$\text{2}^{+}$}, \textcolor{colorLKHRIPT}{$\text{3}^{+}$} & 0.00 & 0.93 & 0.16 &  & 0.00 & 0.99 & 0.07 & \textcolor{colorEAXRGPX}{$\text{2}^{+}$}\\

\cmidrule{3-16}
 & 1000 & $0.00001$ & 10 & $0.00$ & \cellcolor{gray!20}{\textbf{0.89}} & 0.18 & \textcolor{colorEAXRGPX}{$\text{2}^{+}$} & 0.00 & 0.16 & 0.20 &  & 0.00 & 0.84 & 0.27 & \textcolor{colorEAXRGPX}{$\text{2}^{+}$}\\

 & 1000 & $0.00001$ & 50 & $0.00$ & \cellcolor{gray!20}{\textbf{1.00}} & 0.02 & \textcolor{colorEAXRGPX}{$\text{2}^{+}$}, \textcolor{colorLKHRIPT}{$\text{3}^{+}$} & 0.00 & 0.66 & 0.32 &  & 0.00 & 0.96 & 0.15 & \textcolor{colorEAXRGPX}{$\text{2}^{+}$}\\

 & 1000 & $0.00001$ & 100 & $0.00$ & \cellcolor{gray!20}{\textbf{1.00}} & 0.00 & \textcolor{colorEAXRGPX}{$\text{2}^{+}$}, \textcolor{colorLKHRIPT}{$\text{3}^{+}$} & 0.00 & 0.81 & 0.26 &  & 0.00 & 0.97 & 0.11 & \textcolor{colorEAXRGPX}{$\text{2}^{+}$}\\

\cmidrule{3-16}
 & 1000 & $0.00000$ & 10 & $0.00$ & \cellcolor{gray!20}{\textbf{0.88}} & 0.19 & \textcolor{colorEAXRGPX}{$\text{2}^{+}$} & 0.00 & 0.15 & 0.20 &  & 0.00 & 0.83 & 0.27 & \textcolor{colorEAXRGPX}{$\text{2}^{+}$}\\

 & 1000 & $0.00000$ & 50 & $0.00$ & \cellcolor{gray!20}{\textbf{1.00}} & 0.03 & \textcolor{colorEAXRGPX}{$\text{2}^{+}$}, \textcolor{colorLKHRIPT}{$\text{3}^{+}$} & 0.00 & 0.64 & 0.32 &  & 0.00 & 0.96 & 0.15 & \textcolor{colorEAXRGPX}{$\text{2}^{+}$}\\

 & 1000 & $0.00000$ & 100 & $0.00$ & \cellcolor{gray!20}{\textbf{1.00}} & 0.00 & \textcolor{colorEAXRGPX}{$\text{2}^{+}$}, \textcolor{colorLKHRIPT}{$\text{3}^{+}$} & 0.00 & 0.79 & 0.27 &  & 0.00 & 0.97 & 0.11 & \textcolor{colorEAXRGPX}{$\text{2}^{+}$}\\

\cmidrule{2-16}
 & 2000 & $0.00100$ & 10 & $0.03$ & 0.43 & 0.17 & \textcolor{colorEAXRGPX}{$\text{2}^{+}$} & 0.04 & 0.00 & 0.00 &  & 0.00 & \cellcolor{gray!20}{\textbf{1.00}} & 0.00 & \textcolor{colorEAXR}{$\text{1}^{+}$}, \textcolor{colorEAXRGPX}{$\text{2}^{+}$}\\

 & 2000 & $0.00100$ & 50 & $0.00$ & \cellcolor{gray!20}{\textbf{1.00}} & 0.00 & \textcolor{colorEAXRGPX}{$\text{2}^{+}$} & 0.00 & 0.05 & 0.09 &  & 0.00 & \cellcolor{gray!20}{\textbf{1.00}} & 0.00 & \textcolor{colorEAXRGPX}{$\text{2}^{+}$}\\

 & 2000 & $0.00100$ & 100 & $0.00$ & \cellcolor{gray!20}{\textbf{1.00}} & 0.00 & \textcolor{colorEAXRGPX}{$\text{2}^{+}$} & 0.00 & 0.49 & 0.21 &  & 0.00 & \cellcolor{gray!20}{\textbf{1.00}} & 0.00 & \textcolor{colorEAXRGPX}{$\text{2}^{+}$}\\

\cmidrule{3-16}
 & 2000 & $0.00010$ & 10 & $0.03$ & 0.18 & 0.16 & \textcolor{colorEAXRGPX}{$\text{2}^{+}$} & 0.04 & 0.00 & 0.00 &  & 0.00 & \cellcolor{gray!20}{\textbf{0.59}} & 0.30 & \textcolor{colorEAXR}{$\text{1}^{+}$}, \textcolor{colorEAXRGPX}{$\text{2}^{+}$}\\

 & 2000 & $0.00010$ & 50 & $0.00$ & \cellcolor{gray!20}{\textbf{0.99}} & 0.05 & \textcolor{colorEAXRGPX}{$\text{2}^{+}$}, \textcolor{colorLKHRIPT}{$\text{3}^{+}$} & 0.00 & 0.00 & 0.00 &  & 0.00 & 0.89 & 0.18 & \textcolor{colorEAXRGPX}{$\text{2}^{+}$}\\

 & 2000 & $0.00010$ & 100 & $0.00$ & \cellcolor{gray!20}{\textbf{1.00}} & 0.01 & \textcolor{colorEAXRGPX}{$\text{2}^{+}$}, \textcolor{colorLKHRIPT}{$\text{3}^{+}$} & 0.00 & 0.01 & 0.03 &  & 0.00 & 0.95 & 0.12 & \textcolor{colorEAXRGPX}{$\text{2}^{+}$}\\

\cmidrule{3-16}
 & 2000 & $0.00001$ & 10 & $0.03$ & 0.06 & 0.10 & \textcolor{colorEAXRGPX}{$\text{2}^{+}$} & 0.04 & 0.00 & 0.00 &  & 0.00 & \cellcolor{gray!20}{\textbf{0.34}} & 0.28 & \textcolor{colorEAXR}{$\text{1}^{+}$}, \textcolor{colorEAXRGPX}{$\text{2}^{+}$}\\

 & 2000 & $0.00001$ & 50 & $0.00$ & \cellcolor{gray!20}{\textbf{0.83}} & 0.23 & \textcolor{colorEAXRGPX}{$\text{2}^{+}$}, \textcolor{colorLKHRIPT}{$\text{3}^{+}$} & 0.00 & 0.00 & 0.00 &  & 0.00 & 0.71 & 0.32 & \textcolor{colorEAXRGPX}{$\text{2}^{+}$}\\

 & 2000 & $0.00001$ & 100 & $0.00$ & \cellcolor{gray!20}{\textbf{0.94}} & 0.13 & \textcolor{colorEAXRGPX}{$\text{2}^{+}$}, \textcolor{colorLKHRIPT}{$\text{3}^{+}$} & 0.00 & 0.00 & 0.01 &  & 0.00 & 0.83 & 0.28 & \textcolor{colorEAXRGPX}{$\text{2}^{+}$}\\

\cmidrule{3-16}
 & 2000 & $0.00000$ & 10 & $0.03$ & 0.04 & 0.08 & \textcolor{colorEAXRGPX}{$\text{2}^{+}$} & 0.04 & 0.00 & 0.00 &  & 0.00 & \cellcolor{gray!20}{\textbf{0.31}} & 0.27 & \textcolor{colorEAXR}{$\text{1}^{+}$}, \textcolor{colorEAXRGPX}{$\text{2}^{+}$}\\

 & 2000 & $0.00000$ & 50 & $0.00$ & \cellcolor{gray!20}{\textbf{0.78}} & 0.25 & \textcolor{colorEAXRGPX}{$\text{2}^{+}$}, \textcolor{colorLKHRIPT}{$\text{3}^{+}$} & 0.00 & 0.00 & 0.00 &  & 0.00 & 0.67 & 0.32 & \textcolor{colorEAXRGPX}{$\text{2}^{+}$}\\

\multirow{-40}{*}{\raggedright\arraybackslash \textbf{rue}} & 2000 & $0.00000$ & 100 & $0.00$ & \cellcolor{gray!20}{\textbf{0.91}} & 0.17 & \textcolor{colorEAXRGPX}{$\text{2}^{+}$}, \textcolor{colorLKHRIPT}{$\text{3}^{+}$} & 0.00 & 0.00 & 0.01 &  & 0.00 & 0.80 & 0.29 & \textcolor{colorEAXRGPX}{$\text{2}^{+}$}\\
\cmidrule{1-16}
 & 1000 & $0.00100$ & 10 & $0.00$ & \cellcolor{gray!20}{\textbf{1.00}} & 0.00 & \textcolor{colorEAXRGPX}{$\text{2}^{+}$}, \textcolor{colorLKHRIPT}{$\text{3}^{+}$} & 0.00 & 0.99 & 0.06 &  & 0.03 & 0.97 & 0.11 & \\

 & 1000 & $0.00100$ & 50 & $0.00$ & \cellcolor{gray!20}{\textbf{1.00}} & 0.00 & \textcolor{colorLKHRIPT}{$\text{3}^{+}$} & 0.00 & \cellcolor{gray!20}{\textbf{1.00}} & 0.00 & \textcolor{colorLKHRIPT}{$\text{3}^{+}$} & 0.00 & 0.99 & 0.06 & \\

 & 1000 & $0.00100$ & 100 & $0.00$ & \cellcolor{gray!20}{\textbf{1.00}} & 0.00 & \textcolor{colorLKHRIPT}{$\text{3}^{+}$} & 0.00 & \cellcolor{gray!20}{\textbf{1.00}} & 0.00 & \textcolor{colorLKHRIPT}{$\text{3}^{+}$} & 0.00 & 1.00 & 0.04 & \\

\cmidrule{3-16}
 & 1000 & $0.00010$ & 10 & $0.00$ & \cellcolor{gray!20}{\textbf{0.99}} & 0.07 & \textcolor{colorEAXRGPX}{$\text{2}^{+}$}, \textcolor{colorLKHRIPT}{$\text{3}^{+}$} & 0.00 & 0.66 & 0.31 &  & 0.03 & 0.80 & 0.29 & \textcolor{colorEAXRGPX}{$\text{2}^{+}$}\\

 & 1000 & $0.00010$ & 50 & $0.00$ & \cellcolor{gray!20}{\textbf{1.00}} & 0.01 & \textcolor{colorEAXRGPX}{$\text{2}^{+}$}, \textcolor{colorLKHRIPT}{$\text{3}^{+}$} & 0.00 & 0.96 & 0.13 &  & 0.00 & 0.94 & 0.17 & \\

 & 1000 & $0.00010$ & 100 & $0.00$ & \cellcolor{gray!20}{\textbf{1.00}} & 0.01 & \textcolor{colorEAXRGPX}{$\text{2}^{+}$}, \textcolor{colorLKHRIPT}{$\text{3}^{+}$} & 0.00 & 0.98 & 0.09 & \textcolor{colorLKHRIPT}{$\text{3}^{+}$} & 0.00 & 0.97 & 0.13 & \\

\cmidrule{3-16}
 & 1000 & $0.00001$ & 10 & $0.00$ & \cellcolor{gray!20}{\textbf{0.95}} & 0.14 & \textcolor{colorEAXRGPX}{$\text{2}^{+}$}, \textcolor{colorLKHRIPT}{$\text{3}^{+}$} & 0.00 & 0.48 & 0.34 &  & 0.03 & 0.68 & 0.35 & \textcolor{colorEAXRGPX}{$\text{2}^{+}$}\\

 & 1000 & $0.00001$ & 50 & $0.00$ & \cellcolor{gray!20}{\textbf{1.00}} & 0.03 & \textcolor{colorEAXRGPX}{$\text{2}^{+}$}, \textcolor{colorLKHRIPT}{$\text{3}^{+}$} & 0.00 & 0.88 & 0.22 &  & 0.00 & 0.87 & 0.26 & \\

 & 1000 & $0.00001$ & 100 & $0.00$ & \cellcolor{gray!20}{\textbf{1.00}} & 0.03 & \textcolor{colorEAXRGPX}{$\text{2}^{+}$}, \textcolor{colorLKHRIPT}{$\text{3}^{+}$} & 0.00 & 0.94 & 0.17 &  & 0.00 & 0.92 & 0.22 & \\

\cmidrule{3-16}
 & 1000 & $0.00000$ & 10 & $0.00$ & \cellcolor{gray!20}{\textbf{0.94}} & 0.15 & \textcolor{colorEAXRGPX}{$\text{2}^{+}$}, \textcolor{colorLKHRIPT}{$\text{3}^{+}$} & 0.00 & 0.45 & 0.34 &  & 0.03 & 0.65 & 0.36 & \textcolor{colorEAXRGPX}{$\text{2}^{+}$}\\

 & 1000 & $0.00000$ & 50 & $0.00$ & \cellcolor{gray!20}{\textbf{1.00}} & 0.04 & \textcolor{colorEAXRGPX}{$\text{2}^{+}$}, \textcolor{colorLKHRIPT}{$\text{3}^{+}$} & 0.00 & 0.87 & 0.24 &  & 0.00 & 0.86 & 0.27 & \\

 & 1000 & $0.00000$ & 100 & $0.00$ & \cellcolor{gray!20}{\textbf{1.00}} & 0.03 & \textcolor{colorEAXRGPX}{$\text{2}^{+}$}, \textcolor{colorLKHRIPT}{$\text{3}^{+}$} & 0.00 & 0.93 & 0.18 &  & 0.00 & 0.91 & 0.23 & \\

\cmidrule{2-16}
 & 2000 & $0.00100$ & 10 & $0.02$ & 0.78 & 0.26 & \textcolor{colorEAXRGPX}{$\text{2}^{+}$} & 0.06 & 0.00 & 0.00 &  & 0.09 & \cellcolor{gray!20}{\textbf{0.89}} & 0.21 & \textcolor{colorEAXR}{$\text{1}^{+}$}, \textcolor{colorEAXRGPX}{$\text{2}^{+}$}\\

 & 2000 & $0.00100$ & 50 & $0.00$ & \cellcolor{gray!20}{\textbf{1.00}} & 0.00 & \textcolor{colorEAXRGPX}{$\text{2}^{+}$}, \textcolor{colorLKHRIPT}{$\text{3}^{+}$} & 0.01 & 0.45 & 0.35 &  & 0.04 & 0.99 & 0.06 & \textcolor{colorEAXRGPX}{$\text{2}^{+}$}\\

 & 2000 & $0.00100$ & 100 & $0.00$ & \cellcolor{gray!20}{\textbf{1.00}} & 0.00 & \textcolor{colorEAXRGPX}{$\text{2}^{+}$}, \textcolor{colorLKHRIPT}{$\text{3}^{+}$} & 0.01 & 0.87 & 0.19 &  & 0.04 & 1.00 & 0.03 & \textcolor{colorEAXRGPX}{$\text{2}^{+}$}\\

\cmidrule{3-16}
 & 2000 & $0.00010$ & 10 & $0.02$ & \cellcolor{gray!20}{\textbf{0.51}} & 0.28 & \textcolor{colorEAXRGPX}{$\text{2}^{+}$}, \textcolor{colorLKHRIPT}{$\text{3}^{+}$} & 0.06 & 0.00 & 0.00 &  & 0.09 & 0.40 & 0.33 & \textcolor{colorEAXRGPX}{$\text{2}^{+}$}\\

 & 2000 & $0.00010$ & 50 & $0.00$ & \cellcolor{gray!20}{\textbf{0.99}} & 0.06 & \textcolor{colorEAXRGPX}{$\text{2}^{+}$}, \textcolor{colorLKHRIPT}{$\text{3}^{+}$} & 0.01 & 0.03 & 0.10 &  & 0.04 & 0.80 & 0.29 & \textcolor{colorEAXRGPX}{$\text{2}^{+}$}\\

 & 2000 & $0.00010$ & 100 & $0.00$ & \cellcolor{gray!20}{\textbf{1.00}} & 0.02 & \textcolor{colorEAXRGPX}{$\text{2}^{+}$}, \textcolor{colorLKHRIPT}{$\text{3}^{+}$} & 0.01 & 0.15 & 0.18 &  & 0.04 & 0.88 & 0.24 & \textcolor{colorEAXRGPX}{$\text{2}^{+}$}\\

\cmidrule{3-16}
 & 2000 & $0.00001$ & 10 & $0.02$ & \cellcolor{gray!20}{\textbf{0.32}} & 0.26 & \textcolor{colorEAXRGPX}{$\text{2}^{+}$}, \textcolor{colorLKHRIPT}{$\text{3}^{+}$} & 0.06 & 0.00 & 0.00 &  & 0.09 & 0.19 & 0.25 & \textcolor{colorEAXRGPX}{$\text{2}^{+}$}\\

 & 2000 & $0.00001$ & 50 & $0.00$ & \cellcolor{gray!20}{\textbf{0.93}} & 0.16 & \textcolor{colorEAXRGPX}{$\text{2}^{+}$}, \textcolor{colorLKHRIPT}{$\text{3}^{+}$} & 0.01 & 0.01 & 0.05 &  & 0.04 & 0.60 & 0.35 & \textcolor{colorEAXRGPX}{$\text{2}^{+}$}\\

 & 2000 & $0.00001$ & 100 & $0.00$ & \cellcolor{gray!20}{\textbf{0.98}} & 0.09 & \textcolor{colorEAXRGPX}{$\text{2}^{+}$}, \textcolor{colorLKHRIPT}{$\text{3}^{+}$} & 0.01 & 0.04 & 0.10 &  & 0.04 & 0.72 & 0.34 & \textcolor{colorEAXRGPX}{$\text{2}^{+}$}\\

\cmidrule{3-16}
 & 2000 & $0.00000$ & 10 & $0.02$ & \cellcolor{gray!20}{\textbf{0.28}} & 0.25 & \textcolor{colorEAXRGPX}{$\text{2}^{+}$}, \textcolor{colorLKHRIPT}{$\text{3}^{+}$} & 0.06 & 0.00 & 0.00 &  & 0.09 & 0.15 & 0.24 & \textcolor{colorEAXRGPX}{$\text{2}^{+}$}\\

 & 2000 & $0.00000$ & 50 & $0.00$ & \cellcolor{gray!20}{\textbf{0.90}} & 0.18 & \textcolor{colorEAXRGPX}{$\text{2}^{+}$}, \textcolor{colorLKHRIPT}{$\text{3}^{+}$} & 0.01 & 0.01 & 0.05 &  & 0.04 & 0.55 & 0.36 & \textcolor{colorEAXRGPX}{$\text{2}^{+}$}\\

\multirow{-40}{*}{\raggedright\arraybackslash \textbf{structured}} & 2000 & $0.00000$ & 100 & $0.00$ & \cellcolor{gray!20}{\textbf{0.97}} & 0.11 & \textcolor{colorEAXRGPX}{$\text{2}^{+}$}, \textcolor{colorLKHRIPT}{$\text{3}^{+}$} & 0.01 & 0.03 & 0.09 &  & 0.04 & 0.68 & 0.35 & \textcolor{colorEAXRGPX}{$\text{2}^{+}$}\\
\bottomrule
\end{tabular}
\end{scriptsize}
\end{table*}
\end{scriptsize}


\section{Conclusion and Future Work}
\label{sec:conclusion}

Taking an ``anytime perspective'' in TSP solver benchmarking, i.e., addressing research questions R1 and R2 stated in the introduction, results in detailed insights into solver performances together with respective approximation speed and indicates  structural relationships with instance properties. Specifically, our sophisticated evolutionary approach \cite{bossek2019} for generating instances which are very hard for EAX+r and easy for LKH+r proves to perform extremely well reflecting the huge impact of instance properties on problem hardness.

Next steps will incorporate the insights in anytime performance of TSP solvers gained from our empirical study into automated algorithm selection models (ref. to research question R3). Building on existing high-performing approaches conditioned on the necessity of solving the problem to optimality (e.g., \cite{KKBHTLeveragingTSP}) an extension to the anytime scenario would be highly desirable.
For this purpose, several ingredients need to be developed such as instance features characterizing anytime performance. Here, initial work on so-called probing features exists \cite{KKHT2015} and needs to be extended. Moreover, one of course needs an anytime performance indicator which is not straightforward to design as it has to incorporate different aspects of quality and hence is multi-objective in nature. Interesting concepts in the context of automated algorithm configuration \cite{lopez2014automatically} will be tested and improved upon. 

Secondly, the derived insights offer very promising potential in terms of hybridization of inexact TSP solvers as performance rankings differ along the optimization runs with increasing approximation quality. Comparing, e.g., maximum first hitting times, EAX+r generates substantial improvements very fast, is then overtaken by LKH+r (or respective LKH variants) while it clearly is the single-best solver referring to final approximation qualities. Of course, this behaviour is dependent on the instances' structural properties such that a hybridized TSP solver variant that is capable of processing instance features could vastly improve approximation speed and final quality.

\section*{Acknowledgment}
P. Kerschke and H. Trautmann acknowledge support by the European Research Center for Information Systems (ERCIS). J. Bossek was supported by the Australian Research Council (ARC) through grant DP160102401.

\bibliographystyle{unsrt}  
\bibliography{bib}

\end{document}